\documentclass{article}

\usepackage{arxiv}

\usepackage{graphicx}
\usepackage[british]{babel}

\usepackage{xcolor}
\definecolor{newcolor}{rgb}{.8,.349,.1}
\usepackage[linesnumbered,ruled]{algorithm2e}
\usepackage{graphicx}
\usepackage{epstopdf}
\usepackage{subcaption}
\usepackage{mwe}
\usepackage{hyperref}
\usepackage{amsmath}

\usepackage{graphicx}
\usepackage{epstopdf}

\title{DTW-Merge: A Novel Data Augmentation Technique for Time Series Classification}

\author{
	Mohammad Akyash \\
	Electrical Engineering Department\\
	Sharif University of Technology\\
	Tehran, Iran \\
	\texttt{mh\_akyash@yahoo.com} \\
	\And
	Hoda Mohammadzade \\
	Electrical Engineering Department\\
	Sharif University of Technology\\
	Tehran, Iran \\
	\texttt{hoda@sharif.edu} \\
	\And
	Hamid Behroozi \\
	Electrical Engineering Department\\
	Sharif University of Technology\\
	Tehran, Iran \\
	\texttt{behroozi@sharif.edu} \\
}

\begin{document}
	\maketitle
	\begin{abstract}
In recent years, neural networks achieved much success in various applications. The main challenge in training deep neural networks is the lack of sufficient data to improve the model's generalization and avoid overfitting. One of the solutions is to generate new training samples. This paper proposes a novel data augmentation method for time series based on Dynamic Time Warping. This method is inspired by the concept that warped parts of two time series have similar temporal properties and therefore, exchanging them between the two series generates  a new training sample. The proposed method selects an element of the optimal warping path randomly and then exchanges the segments that are aligned together. Exploiting the proposed approach with recently introduced ResNet reveals improved results on the 2018 UCR Time Series Classification Archive. By employing Gradient-weighted Class Activation Mapping (Grad-CAM) and Multidimensional Scaling (MDS), we manifest that our method extract more discriminant features out of time series.

	\end{abstract}


\section{Introduction}
\label{intro}
Time Series Classification (TSC) is a popular research area due to its application in various domains such as human action recognition, biometrics, and vital signs analysis. Since the enormous achievement of deep neural networks in different fields, researchers began to develop neural network based architectures for TSC. One of the earliest networks is Recurrent Neural Networks (RNN), which significantly enhances TSC. Due to computational concerns, this network and its variants such as Long Short-Term Memory (LSTM) are not frequently-used for TSC these days. 

Convolutional Neural Network (CNN) was first introduced for image classification but now play an important role in time series analysis. The objective of these networks is to extract features automatically from inputs. CNN is used for TSC by substitution of conventional 2D filters with 1D ones. Multi-channel CNN (MC-CNN) \cite{Zheng_2015}  is introduced for the classification of multivariate sequences. Multi-Scale CNN (MCNN) \cite{Cui_2016} is proposed for univariate TSC, but due to massive preprocessing on the inputs, it is not considered as the first choice. The LSTM-FCN architecture \cite{Karim_2018} parallels an LSTM and Fully Convolutional Layer (FCN). The authors also suggested dimension shuffling before the LSTM network, which improves the results. In \cite{Wang_2017}, three end-to-end structures are proposed as a baseline for TSC. The first one is Multi-Layer Perceptron (MLP), which uses fully connected layers as a feature extractor. The second is FCN, and the third is the Residual Network (ResNet). The main advantage of these baselines is that they do not need heavy preprocessing; nevertheless, they achieved state-of-the-art results at the time of introduction.

One of the main challenges in training neural networks is the lack of adequate samples, which causes the model to be overfitted, consequently, lower accuracy for unseen samples. One solution is \textit{data augmentation}, which refers to generating new training samples to generalize the model. This technique aims to create samples that assist the model in extracting more discriminant features. Data augmentation is well-established for image classification applications and now is one of the undetachable parts of image analysis. Image augmentation methods include rotating, cropping, and scaling. Data augmentation is yet to be developed specifically for time series, and many of the suggested methods are influenced by the ones for images and might not be suitable for a wide range of applications.

In this paper, a new method for data augmentation is proposed called DTW-Merge. This method is based on Dynamic Time Warping (DTW), which is a well-known tool for aligning time series. We examine the proposed data augmentation method with the ResNet on all datasets of the UCR TSC benchmark \cite{Dau2019}. The experiments show that DTW-Merge could considerably improve the results compared to when no augmentation is used. By utilizing the grad-CAM (Gradient-weighted Class Activation Mapping) \cite{Selvaraju_2019} Multidimensional Scaling (MDS) techniques, we manifest that generated data help the network to learn more discriminative features. We also employ DTW-Merge for 1-Nearest Neighbor DTW (1NN-DTW), which shows no tangible difference in rates.

The rest of the paper is contributed as follows:  In Section\ref{sec:related}, we review some of the previously proposed methods for time series data augmentation; in Section \ref{sec:method}, we revise the DTW algorithm briefly, then propose the DTW-Merge.
%\footnote{Our code and detailed results available on: \url{https://github.com/mhakyash/Merge-DTW}} 
Section \ref{sec:experiments} is dedicated to the experiments and in Section \ref{sec:conclusion}, we conclude the paper.

\section{Related works}
\label{sec:related}

Most of the presented time series data augmentation methods are based on simple transformation, such as Jittering (adding noise to time series), Rotation (Flipping for univariate time series), Magnitude Warping (Warping signal's magnitude), Permutation (Changing position of randomly selected segments), and Scaling (multiplying the magnitude of a sequence by a random number). In \cite{Um_2017}, some of these methods have been evaluated on Parkinson's disease dataset and showed improvement on sequence classification using CNN. Slicing Window \cite{Cui_2016}, firstly, used with MCNN. It operates for time series as same as cropping for images, assuming that cropping sequence would preserve the corresponding label. Window Warping \cite{Guennec_2016} is proposed to reduce the overfitting of CNN and now is one of the most used techniques for data augmentation. This method is based on warping a random window through time. Results reported on \cite{Iwana_2020em} show window warping has the most positive effect among other practical methods for ResNet.  These approaches are based on simple transformation on sequences to generate new samples; however, the main drawback of such methods is that are they able to preserve the samples' labels for all applications.

Some methods take advantage of DTW directly for generating new samples. Suboptimal warped time series generator (SPAWNER) \cite{Kamycki_2020} forces the warping path to pass a random point in distance matrice of Dynamic Time Warping (DTW) and then generate the new time series by averaging the suboptimally aligned sequences. DTW Barycenter Averaging (DBA) \cite{PETITJEAN2011678} is an averaging method for time series, which could be used as a data augmentation method, and in \cite{fawaz2018data}, a weighted version of DBA is evaluated on the UCR archive. Guided Warping \cite{Iwanatime_2020} leverages DTW to mix signals in the time domain. In this method, the optimal DTW warping path is calculated between reference and a teacher sample. Then, the reference's aligned time steps are set to corresponding teacher's ones. Random Guided Warping (RGW) \cite{Iwanatime_2020} and Discriminative Guided Warping (DGW) \cite{Iwanatime_2020} are two variants of this method.

\begin{figure}[!t]
	\centering
	\includegraphics[scale=0.7]{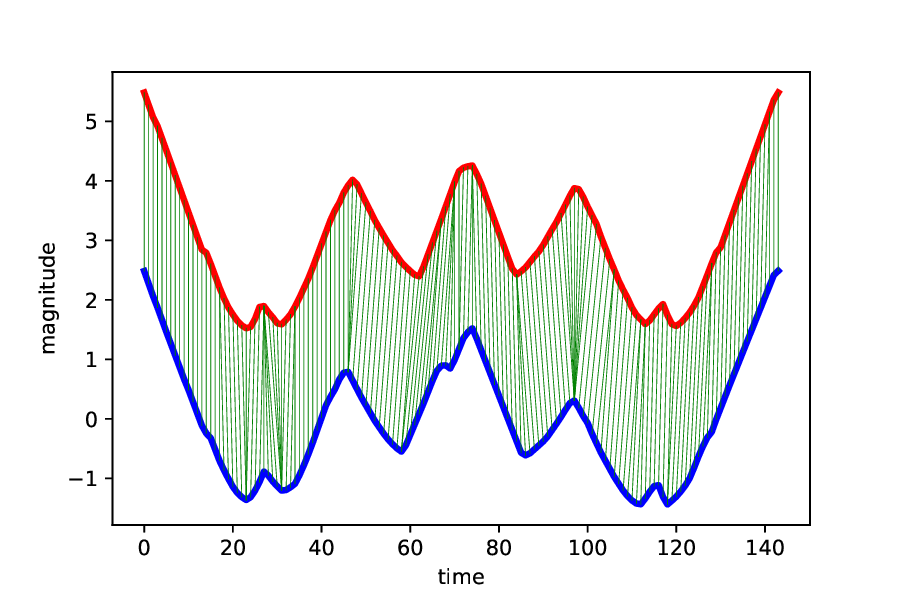}
	\caption{Two samples taken from 'Plane' dataset and corresponding optimal warping path.}
	\label{fig:dtw}
\end{figure}

\section{Method}
\label{sec:method}

DTW is an algorithm for finding similarities of time series and designed to deal with sequences' temporal variations. This technique aims to correspond the time steps of two time series to find the best match between them, called the optimal warping path as depicted in Fig. \ref{fig:dtw}. The DTW distance of the series is reported by the sum of the differences of the aligned samples. The main drawback of DTW is that obtaining the optimal path is time-consuming, especially when the sequences are long because DTW uses the dynamic programming algorithm. There is an approximation of this method called FastDTW \cite{Salvador_2007}, which was suggested to decrease the computational complexity.

DTW is used for TSC for decades and outperforms other similarity measurement techniques such as Euclidean distance because of its ability to handle varying length sequences. One of the common usages of DTW is to employ them in the Nearest Neighbour algorithm. 1NN-DTW tends to assign the label of the most adjacent sample by calculating the DTW distance between the test sequence and all the train ones. Comparably, this method could achieve superior performance in spite of its simplicity.

\begin{figure}[ht]
	\centering
	\begin{minipage}{.9\linewidth}
		\begin{algorithm}[H]
			\SetAlgoLined
			\KwIn{$X$ and $Y$: same class samples}
			\KwOut{$X^{*}$: augmented sample}
			Initialization\;
			Calculating $WarpingPath$ between $X$ and $Y$ using DTW algorithm\; 
			$L$ ← length($WarpingPath$)\;
			$\mu = L/2, \sigma^{2} = L/10$\;		
			r ← a random number from $\mathcal{N}(\mu,\,\sigma^{2})$\;
			$(w_{p,r}, w_{q,r}) = WarpingPath(r)$\;
			$X^{*}_{1}$ ← $X[:w_{p,r}]$, $X^{*}_{2}$ ← $Y[w_{q,r}:]$\;
			$X^{*}$ ← Concat $X^{*}_{1}$ and $X^{*}_{2}$\;
			\caption{DTW-Merge}
			\label{alg:algorithm}
		\end{algorithm}
	\end{minipage}
\end{figure}

\begin{figure*}[]
	%\centering
	
	\includegraphics[scale=0.4]{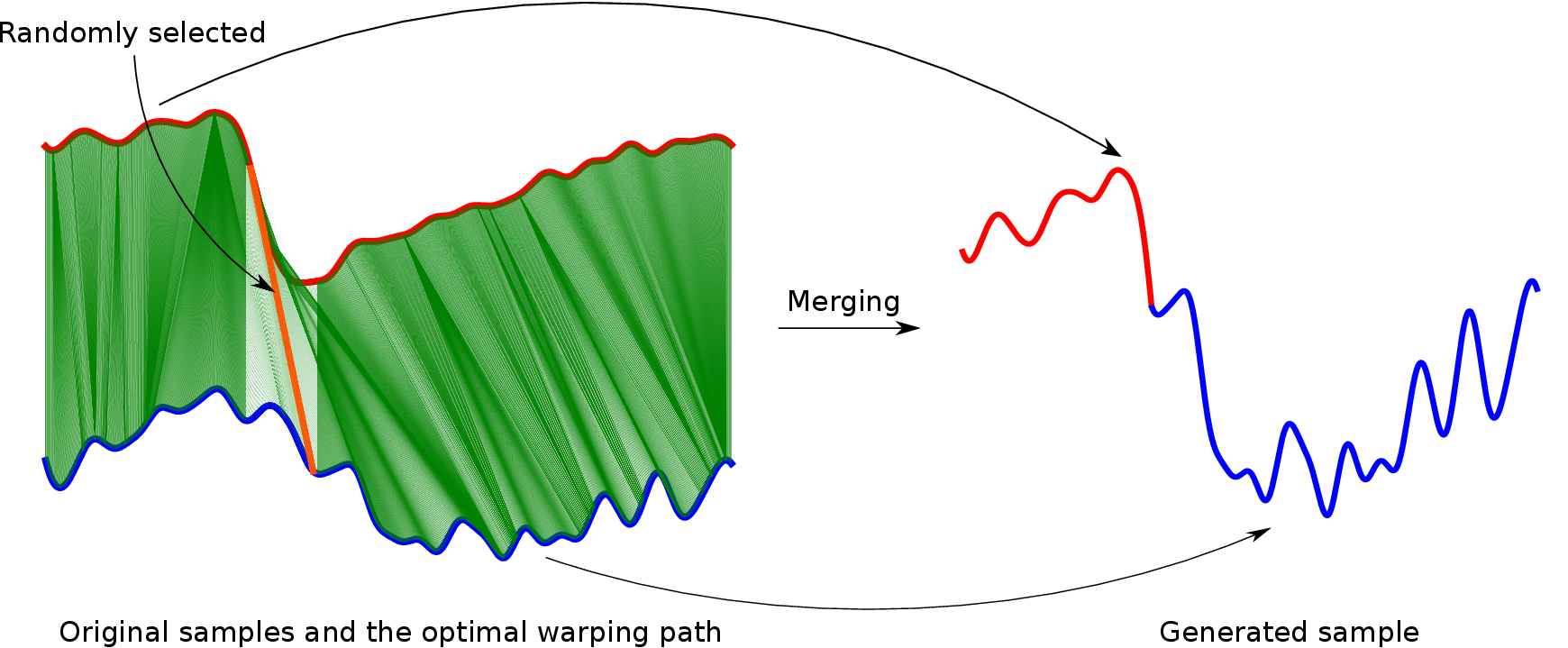}
	\caption{The proposed method. Samples are chosen from 'StarLightCurves' dataset.}
	\label{fig:method}
	
\end{figure*}

In this part, we get through our proposed method which is depicted in Fig. \ref{fig:method}.  Consider $X_{1} : x_{1}, x_{2}, ... x_{M}$ and $Y_{1} : y_{1}, y_{2}, ... y_{N}$ as two sequences from the same class where $x_{j}$ and $y_{j}$ are the $j^{th}$ time steps of the corresponding series. By applying DTW algorithm an optimal warping path is achieved as $W : w_{1}, w_{2}, ..., w_{L}$ where $w_{k} = (w_{p,k}, w_{q, k})$ denotes the aligned samples' indices when $x_{p}$ and $y_{q}$ are matched together. 

Our method is based on exchanging those parts of the time series that have similar temporal properties. We choose these parts using the fact that when two segments of different time series are warped to each other, they have the same temporal characteristics. For choosing these segments, we select a random element of the warping path denoted as $w_{r} = (w_{p,r}, w_{q,r})$ where $r$ is chosen from a Gaussian distribution with the mean of $L/2$ and the variance of $L/10$. Then the augmented sample is generated by merging two parts of the same class sequences as $X^{*} : x_{1}, x_{2}, ..., x_{w_{p,r}}, y_{w_{q,r}}, ... , y_{N}$. By employing this method, the generated sample has identical temporal information, in the dynamic time warping sense, as the original samples. The Algorithm ~\ref{alg:algorithm} shows the steps of the proposed method.

When the time steps of two sequences are aligned via DTW, they have close values and not the same, which causes the connection parts of the time series to be discontinuous. However, this issue does not affect the concept behind the technique, and the generated sequence still has the same temporal features as the original ones. The smoothing filter can correct the discontinuity, but as long as we use the convolution operator, we do not need the smoothing because the 1D filters operate on each sequence and the discontinuous region does not affect the learning procedure.

\begin{figure}[!t]
	\centering
	\includegraphics[scale=0.7]{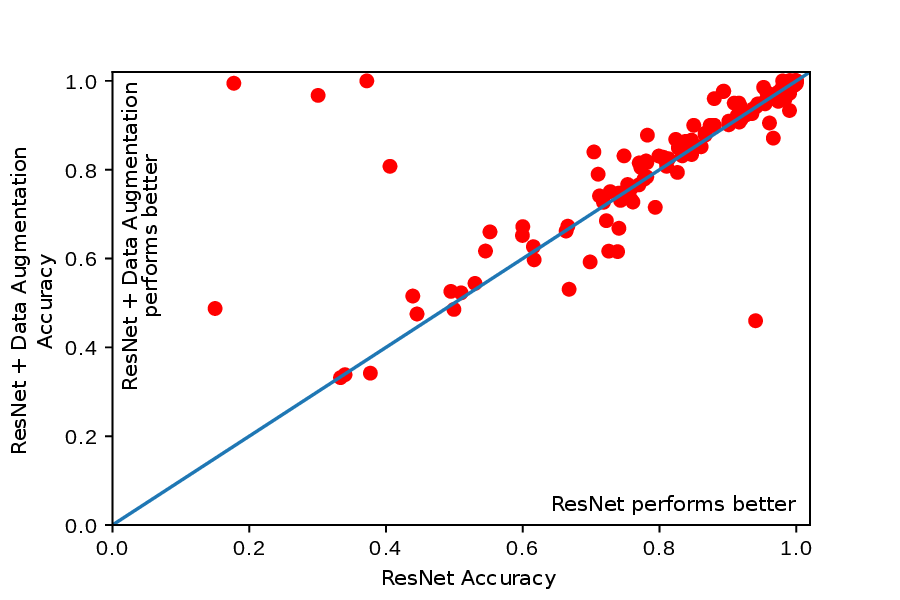}
	\caption{The comparison between ResNet with and without data augmentation on UCR TSC benchmark.}
	\label{fig:overall}
\end{figure}

\section{Experiments}
\label{sec:experiments}

The data augmentation method has been experimented on the UCR TSC benchmark, which contains 128 univariate datasets of various applications \footnote{The implementations and detailed results are available on \url{https://github.com/mhakyash/DTW-Merge}}. Among the networks suggested by \cite{Wang_2017}, we have chosen the ResNet because it is the most complex and deepest network of all three, and we believed that data augmentation could have the most positive effect on this. This network outperforms the FCN in image analysis due to preventing vanishing gradient by flowing gradient through skip connections. We also test the method with 1NN-DTW and will discuss this.

\begin{table}[]
	\centering
	\caption{The performance of ResNet + DTW-Merge. The Win coloumn shows the winning percentage of dataset numbers which this technique outperformed the baseline. The results of the three datasets became a tie.}
	
	\resizebox{.6\textwidth}{!}{%
		\begin{tabular}{cccc}
			\hline
			Type         & \# Datasets & Loss   & Win    \\ \hline
			ECG          & 6           & 50\%   & 50\%   \\
			Hemodynamics & 3           & 0\%    & 100\%  \\
			Power        & 1           & 0\%    & 100\%  \\
			Image        & 32          & 47\%   & 53\%   \\
			EOG          & 2           & 0\%    & 100\%  \\
			Spectro      & 7           & 71\%   & 29\%   \\
			Traffic      & 2           & 0\%    & 100\%  \\
			Sensor       & 28          & 36\%   & 64\%   \\
			HRM          & 1           & 0\%    & 100\%  \\
			Simulated    & 8           & 62.5\% & 37.5\% \\
			Trajectory   & 3           & 100\%  & 0\%    \\
			Device       & 9           & 22\%   & 78\%   \\
			Spectrum     & 4           & 25\%   & 75\%   \\
			EPG          & 2           & 0\%    & 100\%  \\
			Motion       & 17          & 35\%   & 65\%   \\ \hline
		\end{tabular}%
	}
	\label{table:types}
\end{table}

The ResNet architecture includes three stacked blocks; each block contains three convolutional layers with filter sizes of $8$, $5$, and $3$, sequentially. The filter numbers for the convolutional layers of each block are $64$, $128$, and $128$, respectively. After each block, batch normalization and ReLu activation function are placed. After all the blocks, the Global Average Pooling (GAP) layer is used. Glorot Uniform is set as parameters initialization. Adam was used as the optimizer for training the network with $0.001$, $0.9$ and $0.999$ for learning rate, $\beta_{1}$ and $\beta_{2}$, respectively. All datasets were originally split into train and test sets. After augmentation, $25$ percent of training data was considered as the validation set, and the model with minimum validation loss was evaluated on the test set. Every $50$ epochs that no improvement was seen in validation loss, the learning rate was divided by $0.5$ until the minimum of $0.0001$.

Most of the UCR archive datasets are already z-normalized, meaning that they have zero means and unit variances. For the others, we did not apply any preprocessing. Some of the datasets contain different length sequences. Before feeding them to the network, we make their length equal by the following method. First, the average of the length of all the series is calculated; then, if the length of a sequence is greater than the average, we drop the time steps one at a time and randomly until the length becomes equal to the average. On the other hand, if the length is smaller than average, the median of two randomly selected time steps is placed between them. This procedure is also done one at a time until a sequence with the length equal to the average is achieved. By employing this method, we assure that no temporal information is lost.

We compared our results with the ones provided by \cite{Fawaz_2018} on their Github. The authors repeated the training of the ResNet model for five iterations, and we used the mean of all to compare to ours. As it is seen in Fig. \ref{fig:overall} when data augmentation is used, the rates for most datasets are increased. We utilized the paired t-test as hypothesis testing to discover how much the data augmentation improves the ResNet's performance. We obtained the t-value of $2.1$ with a p-value of less than $5\%$ and the average of $2.45\%$ increment for accuracy rates on all datasets. When a data augmentation technique is used, the possibility of overfitting is decreased, and we will be able to use a deeper network. But to make our results comparable to other published papers, we used the three-block ResNet architecture.

Another criterion introduced by \cite{Wang_2017} is Mean Per Class Error (MPCE), which is defined as:

\begin{equation}
PCE_{k} = \frac{Error_{k}}{Number \, of \, Classes}
\end{equation}

\begin{equation}
MPCE = \frac{1}{K} \sum_{k = 1}^{K} PCE_{k}
\label{eq:MPCE}
\end{equation}

Where $PCE$ is calculated for each dataset, and $K$ denotes the number of the datasets. Our method improves this index from $0.0425$ to $0.0365$, which shows better classification accuracy per class for all the datasets.

\begin{figure*}
	\centering
	\begin{subfigure}{0.4\textwidth}
		\centering
		\includegraphics[width=1.2\textwidth]{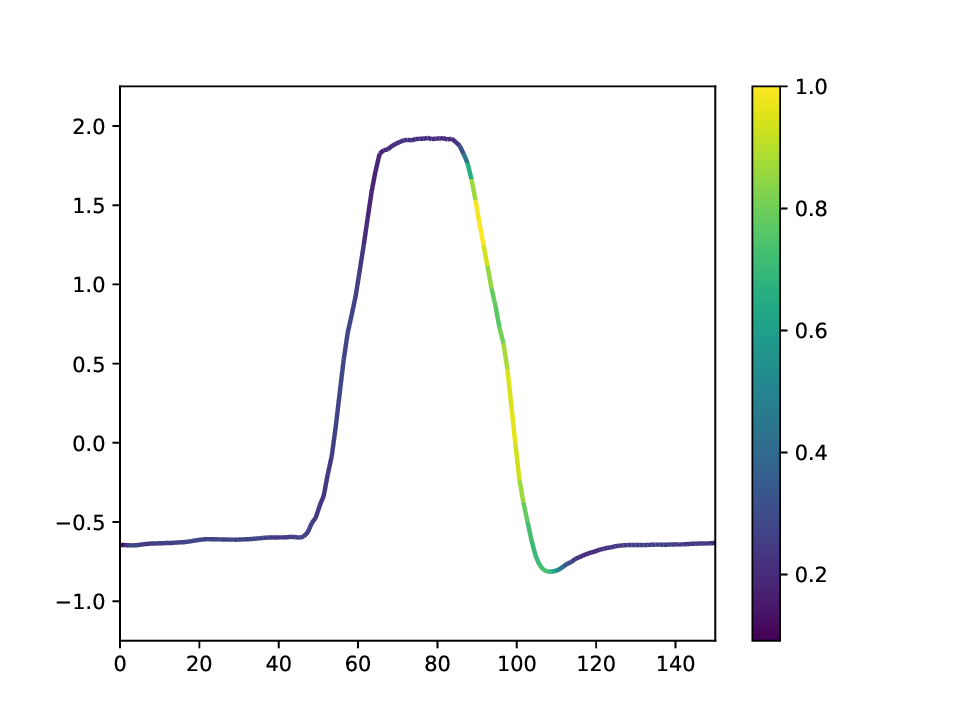}
		\caption[]%
		{{\small A sample of class-1, without data augmentation}}    
		\label{fig:without_1}
	\end{subfigure}
	~ 
	\begin{subfigure}{0.4\textwidth}  
		\centering 
		\includegraphics[width=1.2\textwidth]{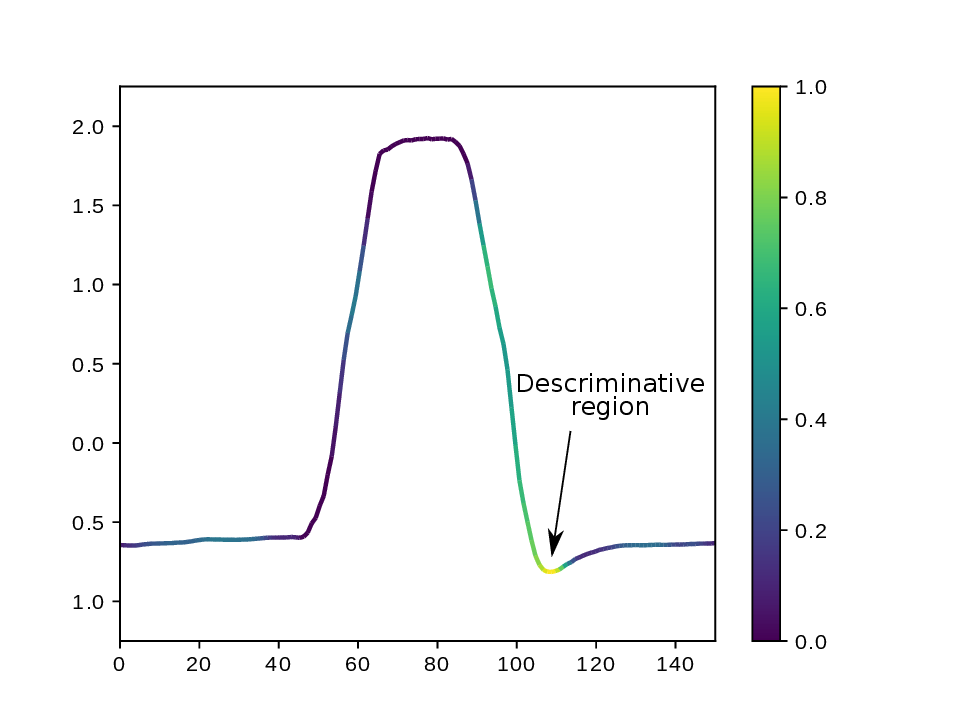}
		\caption[]%
		{{\small A sample of class-1, with data augmentation}}    
		\label{fig:with_1}
	\end{subfigure}
	~ 	
	\begin{subfigure}{0.4\textwidth}   
		\centering 
		\includegraphics[width=1.2\textwidth]{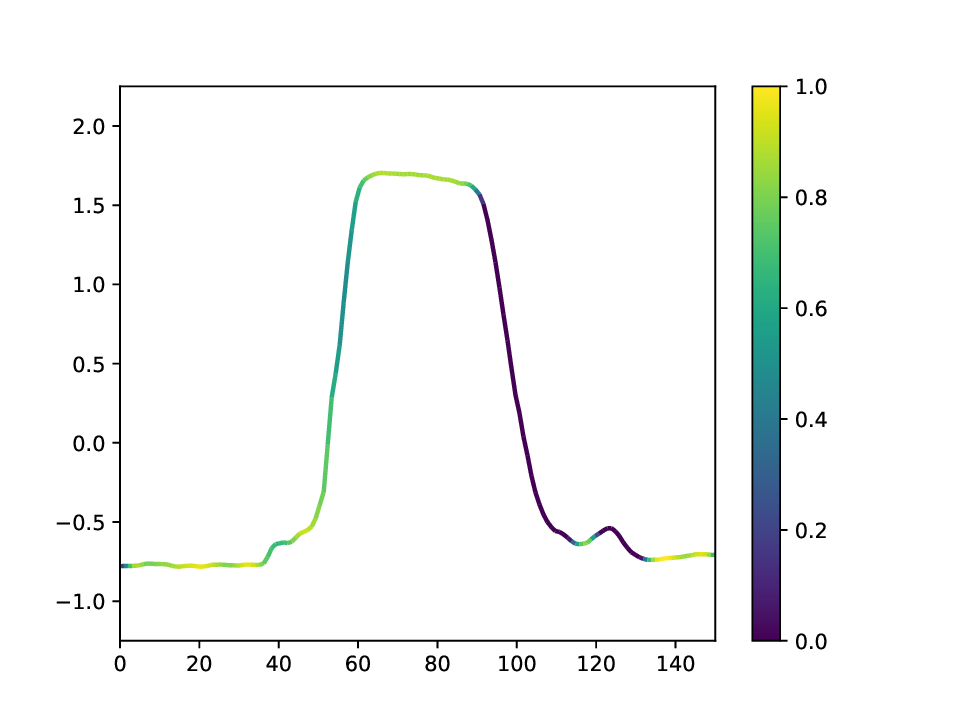}
		\caption[]%
		{{\small A sample of class-2. without data augmentation}}    
		\label{fig:without_2}
	\end{subfigure}
	~ 
	\begin{subfigure}{0.4\textwidth}   
		\centering 
		\includegraphics[width=1.2\textwidth]{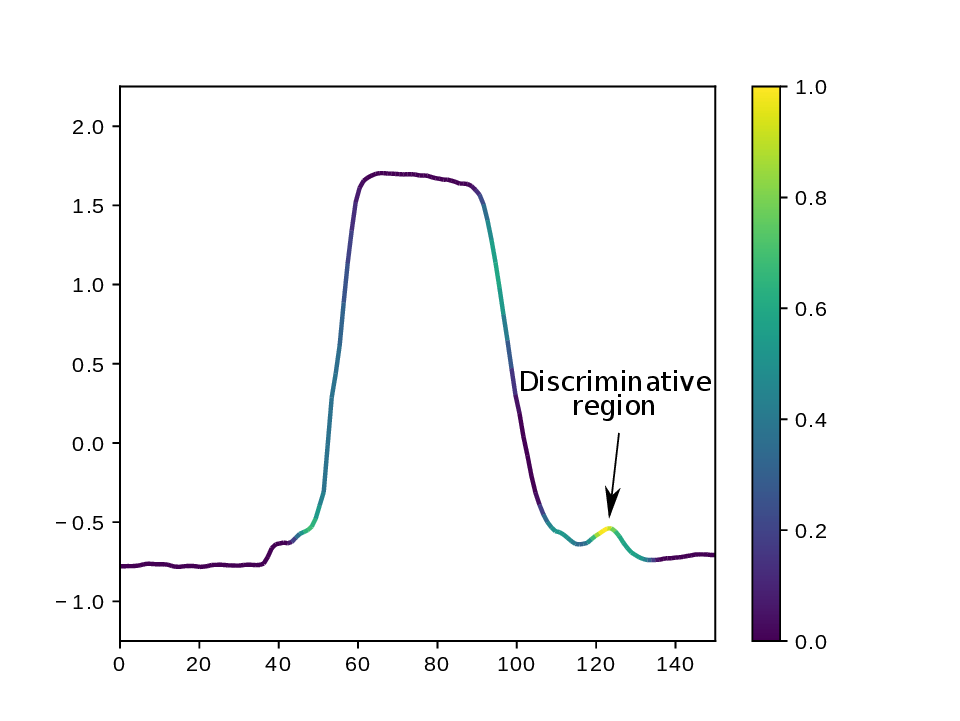}
		\caption[]%
		{{\small A sample of class-2. with data augmentation}}    
		\label{fig:with_2}
	\end{subfigure}
	
	\caption[]
	{\small The utilization of grad-CAM for ResNet with and without data augmentation for GunPoint dataset. The yellow refers to the high contribution, and the dark blue refers to no contribution to the classification. When we use data augmentation, ResNet emphasize on more intensive and discriminative regions. } 
	\label{fig:grad_cam}
	
\end{figure*}

\begin{figure*}[t!]
	\centering
	\begin{subfigure}[t]{0.4\textwidth}
		\centering
		\includegraphics[width=1.2\textwidth]{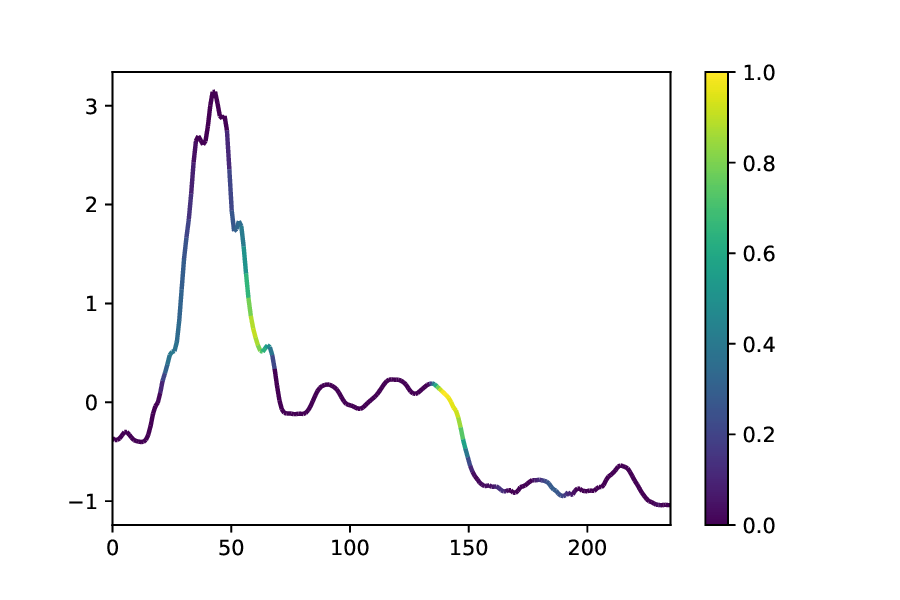}
		\caption[]%
		{{\small A sample of class-1, without data augmentation}}    
		\label{fig:strawberry_without_1}
	\end{subfigure}
	~ 
	\begin{subfigure}[t]{0.4\textwidth}  
		\centering 
		\includegraphics[width=1.2\textwidth]{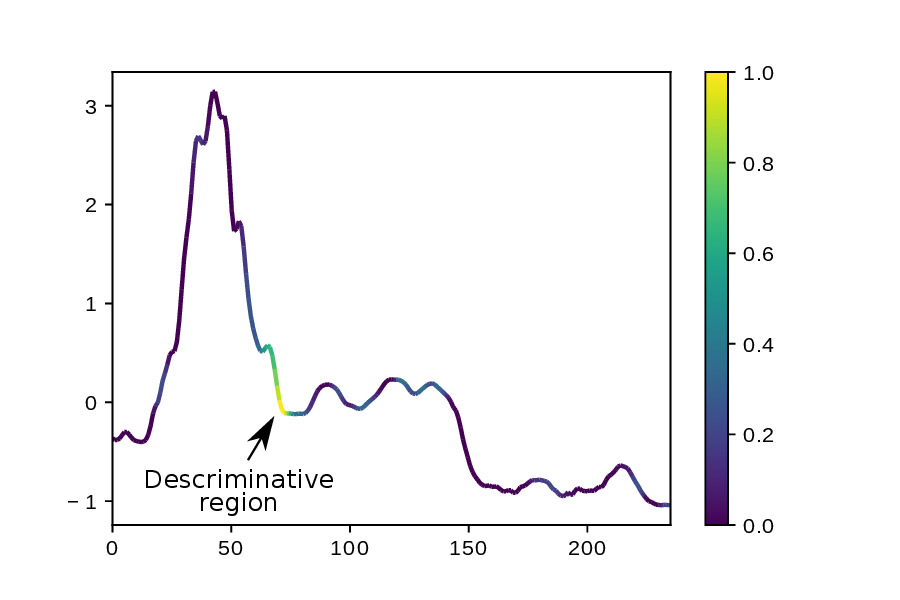}
		\caption[]%
		{{\small A sample of class-1, with data augmentation}}    
		\label{fig:strawberry_with_1}
	\end{subfigure}
	
	\begin{subfigure}[t]{0.4\textwidth}   
		\centering 
		\includegraphics[width=1.2\textwidth]{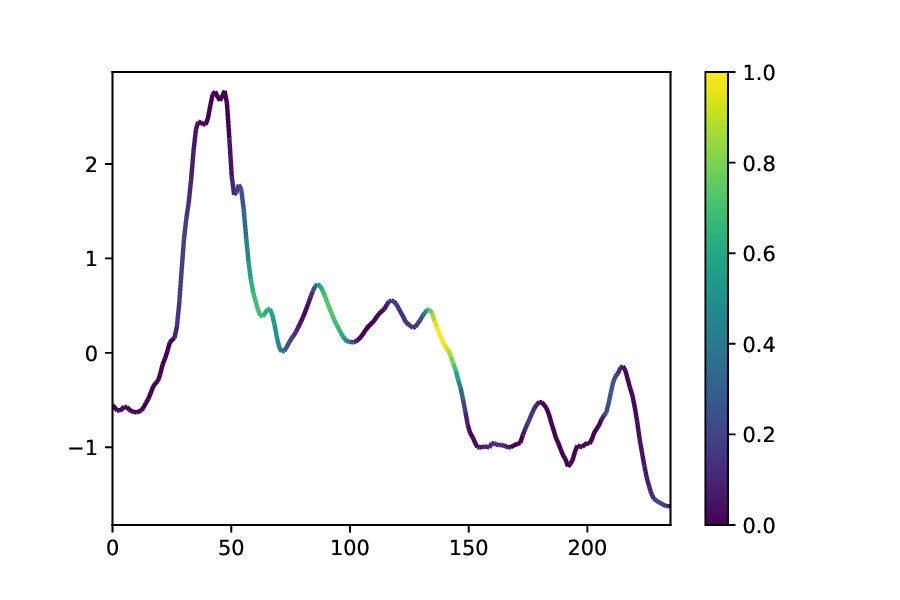}
		\caption[]%
		{{\small A sample of class-2. without data augmentation}}    
		\label{fig:strawberry_without_2}
	\end{subfigure}
	~ 
	\begin{subfigure}[t]{0.4\textwidth}   
		\centering 
		\includegraphics[width=1.2\textwidth]{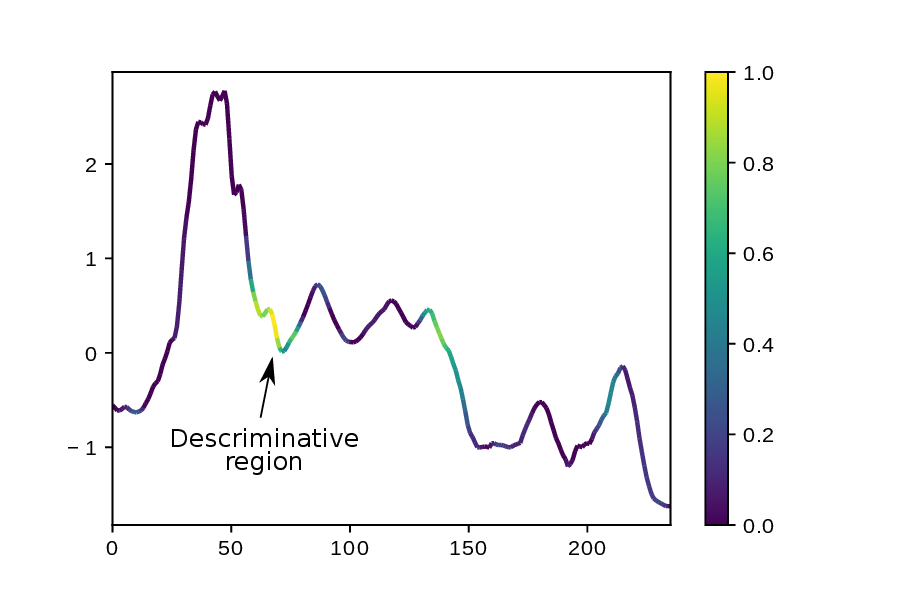}
		\caption[]%
		{{\small A sample of class-2. with data augmentation}}    
		\label{fig:strawberry_with_2}
	\end{subfigure}
	
	\caption[]
	{\small The utilization of grad-CAM for ResNet with and without data augmentation for Strawberry dataset. The yellow refers to the high contribution, and the dark blue refers to no contribution to the classification. When the data augmentation is used, the neural network extract more discriminative features out of each sequence.} 
	\label{fig:grad_cam_2}
	
\end{figure*}

\begin{figure*}
	\centering
	\begin{subfigure}{0.3\textwidth}
		\centering
		\includegraphics[width=1.2\textwidth]{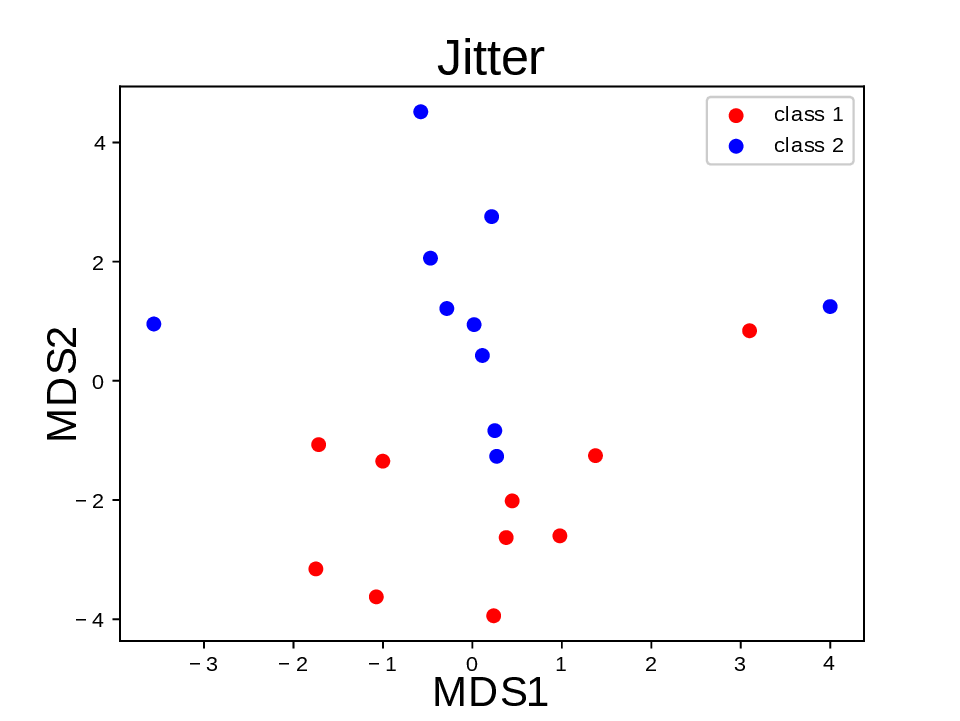}
		\caption[]%
		{{\small Jittering}}    
		\label{fig:MDS_1}
	\end{subfigure}
	~~ 
	\begin{subfigure}{0.3\textwidth}  
		\centering 
		\includegraphics[width=1.2\textwidth]{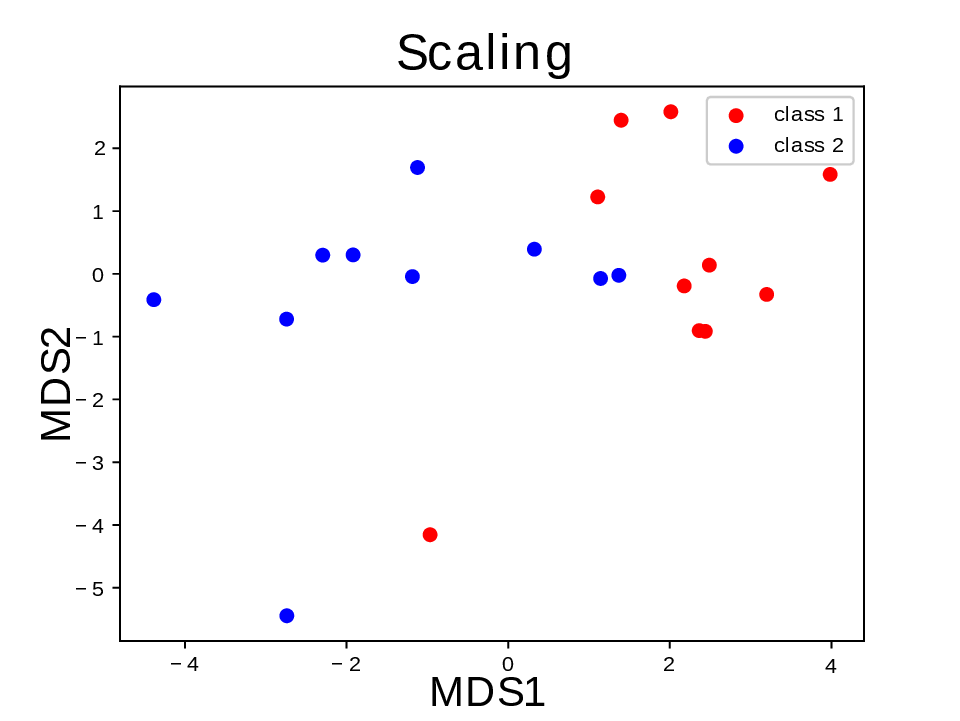}
		\caption[]%
		{{\small Scaling}}    
		\label{fig:MDS_2}
	\end{subfigure}
	~~ 	
	\begin{subfigure}{0.3\textwidth}   
		\centering 
		\includegraphics[width=1.2\textwidth]{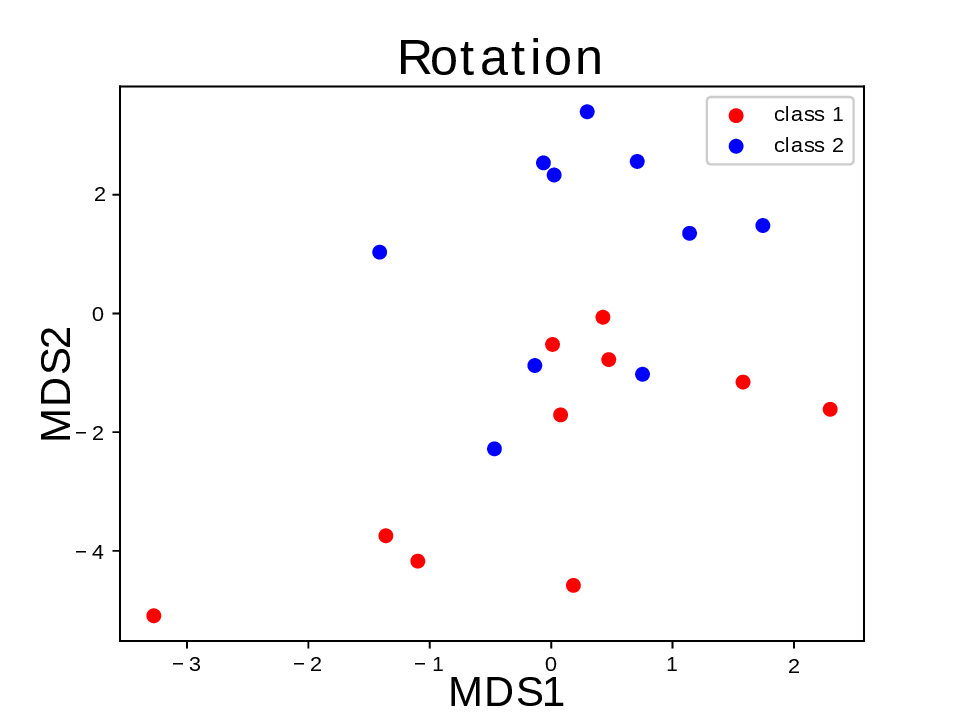}
		\caption[]%
		{{\small Rotation}}    
		\label{fig:MDS_3}
	\end{subfigure}
	\\ 
	\begin{subfigure}{0.3\textwidth}   
		\centering 
		\includegraphics[width=1.2\textwidth]{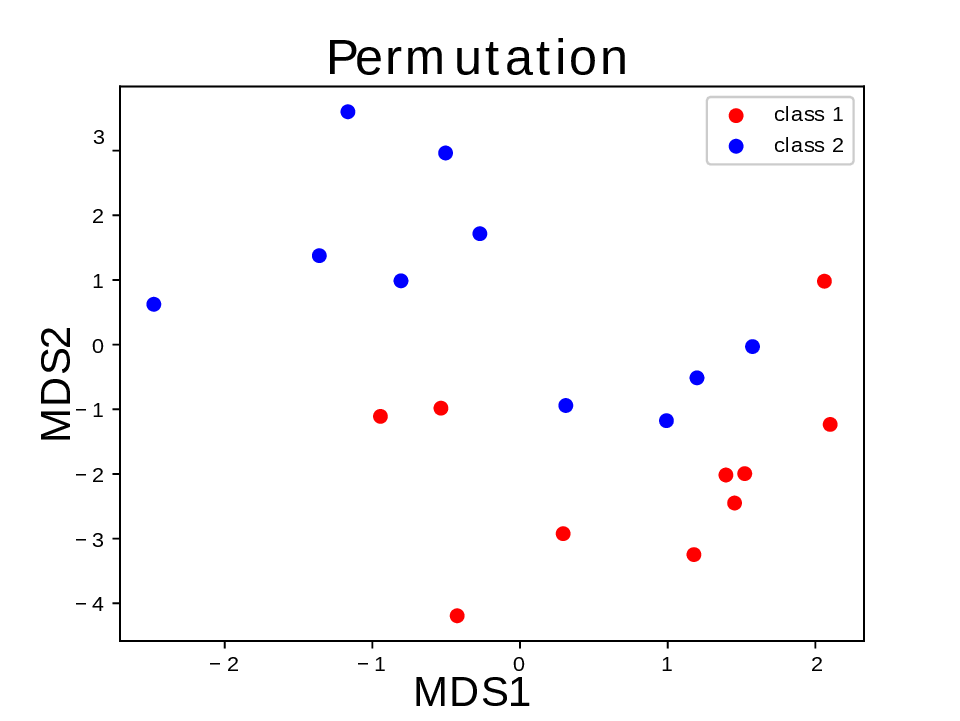}
		\caption[]%
		{{\small Permutation}}    
		\label{fig:MDS_4}
	\end{subfigure}
	~~ 	
	\begin{subfigure}{0.3\textwidth}   
		\centering 
		\includegraphics[width=1.2\textwidth]{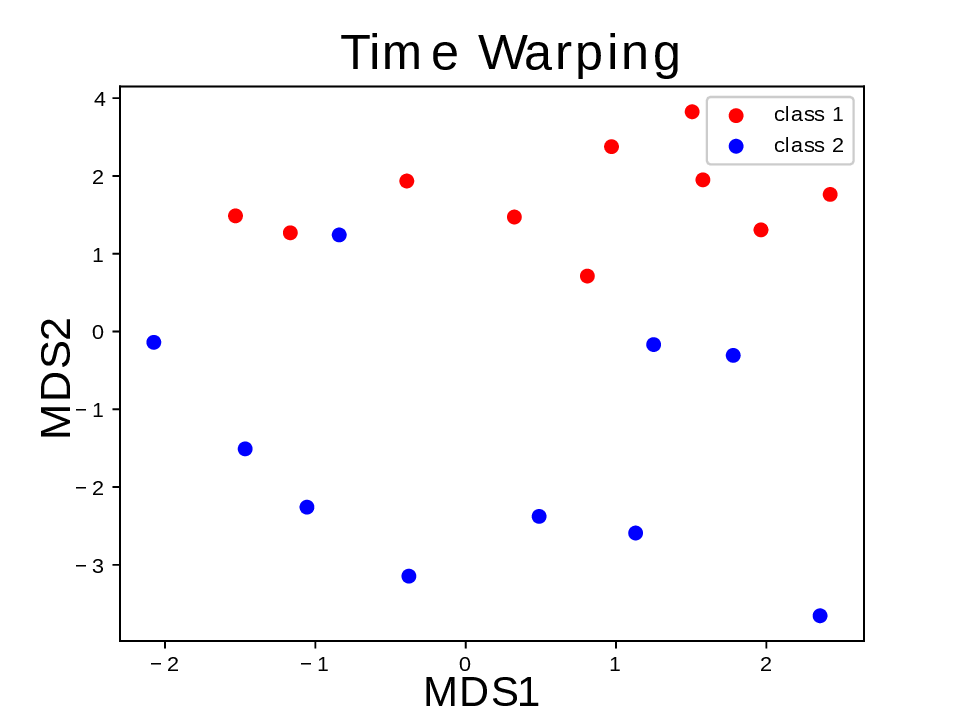}
		\caption[]%
		{{\small Time Warping}}    
		\label{fig:MDS_5}
	\end{subfigure}
	~~ 
	\begin{subfigure}{0.3\textwidth}   
		\centering 
		\includegraphics[width=1.2\textwidth]{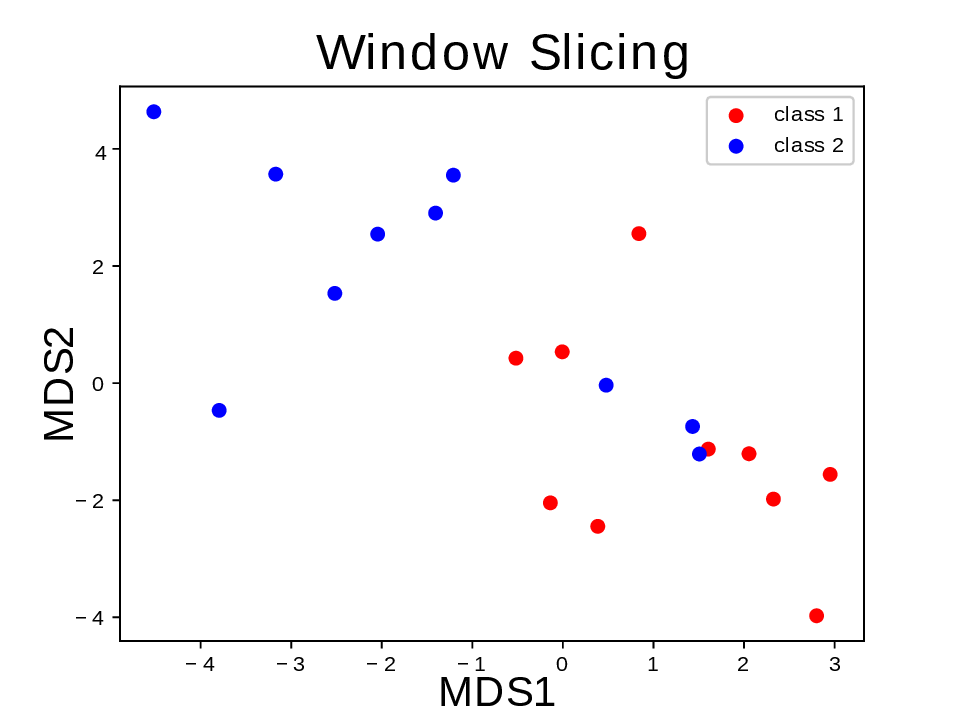}
		\caption[]%
		{{\small Window Slicing}}    
		\label{fig:MDS_6}
	\end{subfigure}
	\\ 
	\begin{subfigure}{0.3\textwidth}   
		\centering 
		\includegraphics[width=1.2\textwidth]{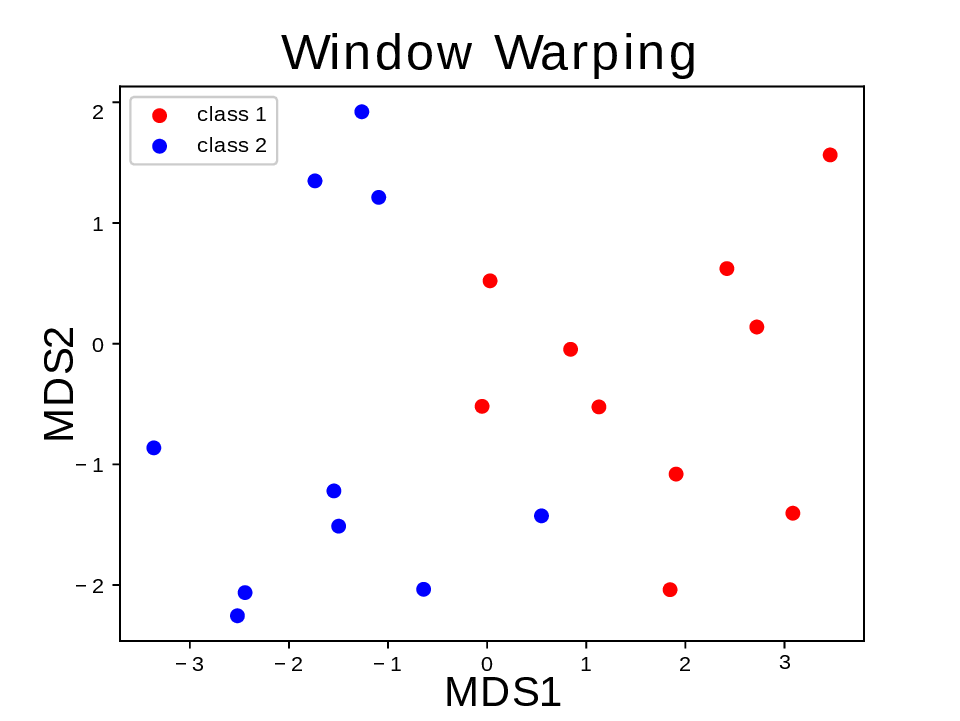}
		\caption[]%
		{{\small Window Warping}}    
		\label{fig:MDS_7}
	\end{subfigure}
	~~ 
	\begin{subfigure}{0.3\textwidth}   
		\centering 
		\includegraphics[width=1.2\textwidth]{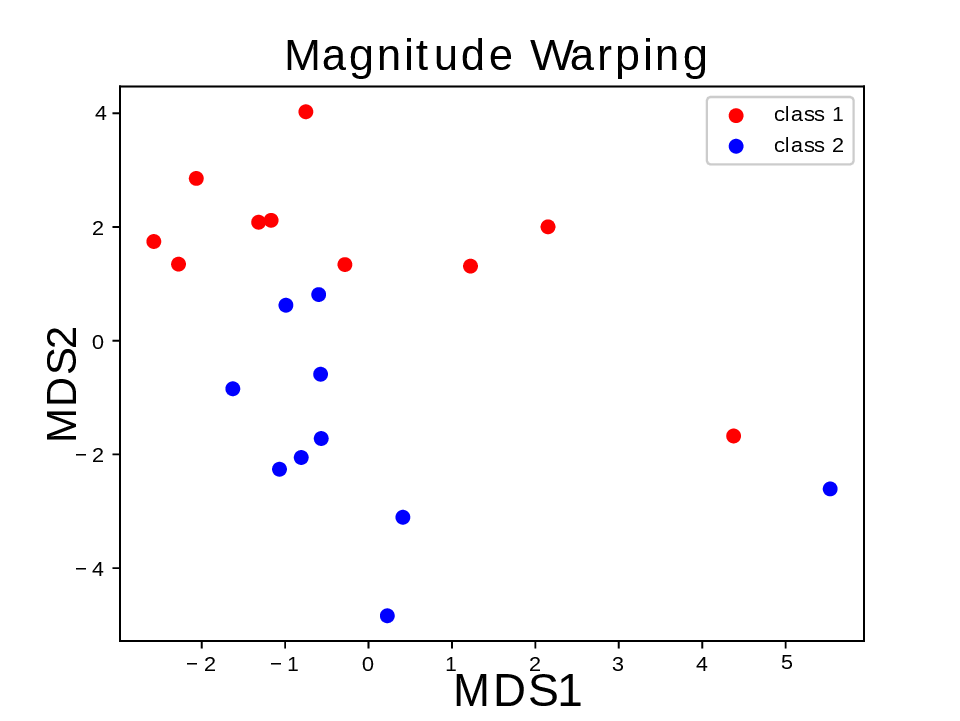}
		\caption[]%
		{{\small Magnitude Warping}}    
		\label{fig:MDS_8}
	\end{subfigure}
	~~
	\begin{subfigure}{0.3\textwidth}   
		\centering 
		\includegraphics[width=1.2\textwidth]{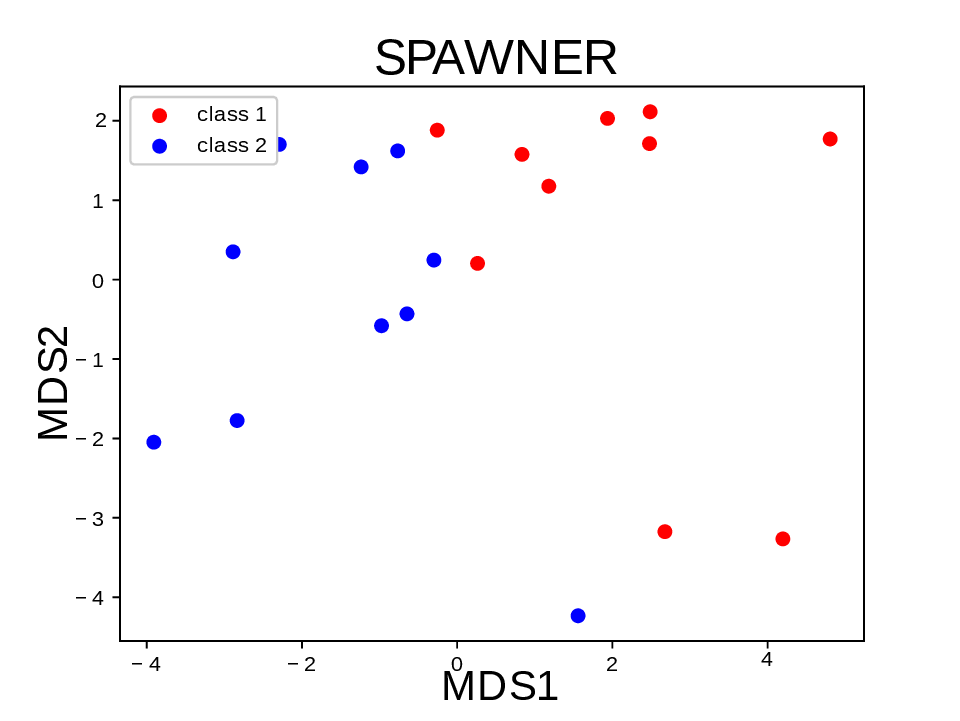}
		\caption[]%
		{{\small SPAWNER}}    
		\label{fig:MDS_11}
	\end{subfigure}
	\\	
	\begin{subfigure}{0.3\textwidth}   
		\centering 
		\includegraphics[width=1.2\textwidth]{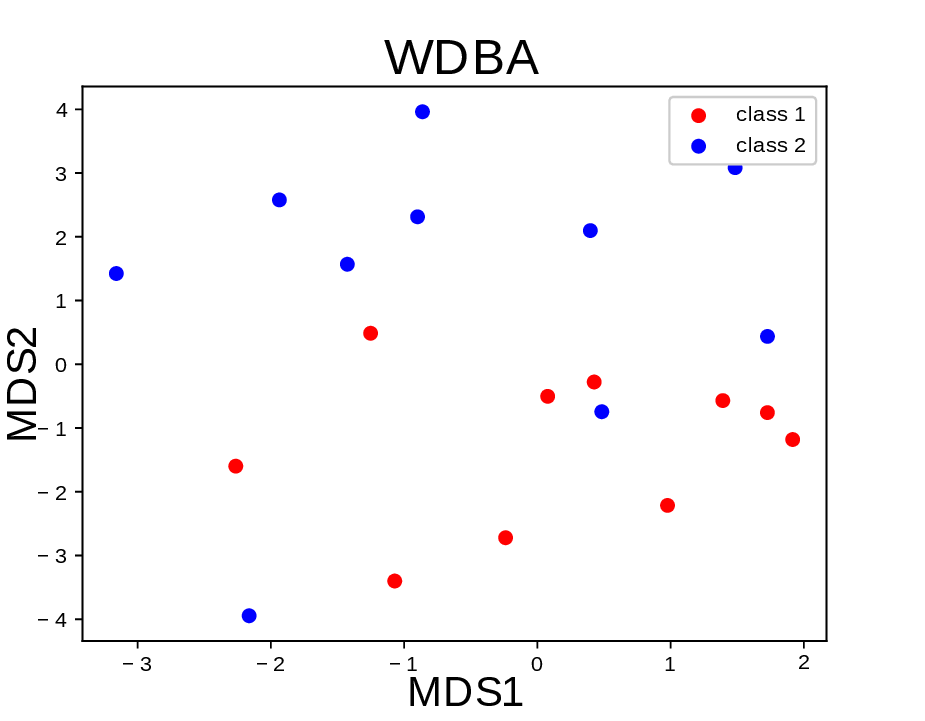}
		\caption[]%
		{{\small WDBA}}    
		\label{fig:MDS_12}
	\end{subfigure}
	~~
	\begin{subfigure}{0.3\textwidth}   
		\centering 
		\includegraphics[width=1.2\textwidth]{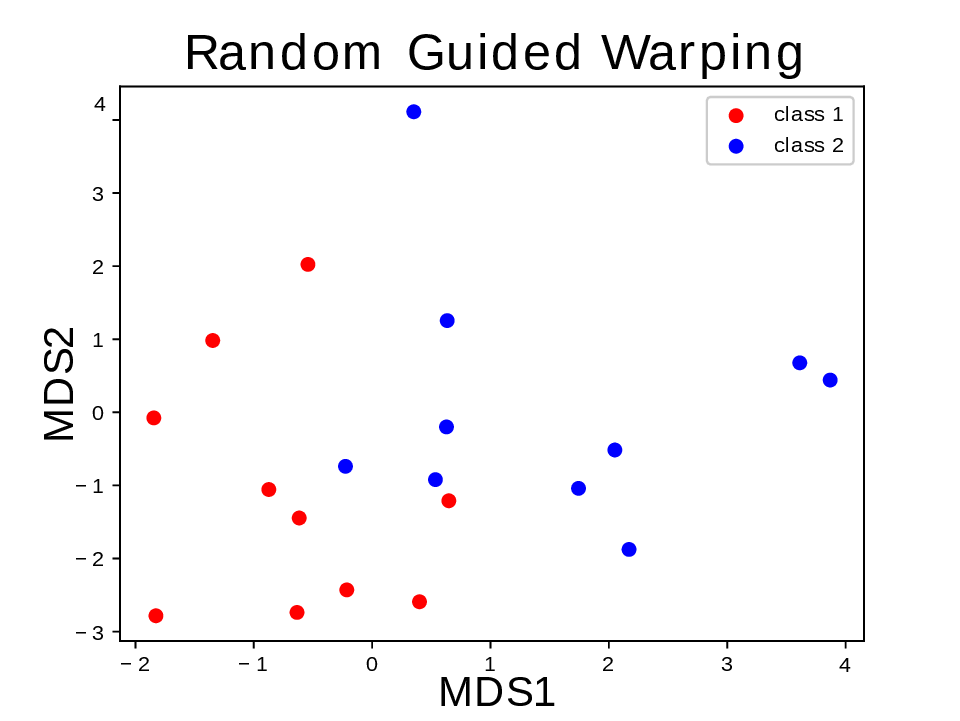}
		\caption[]%
		{{\small RGW}}    
		\label{fig:MDS_13}
	\end{subfigure}
	~~
	%	\begin{subfigure}{0.2\textwidth}   
	%		\centering 
	%		\includegraphics[width=1.2\textwidth]{random_guided_warp_shape.eps}
	%		\caption[]%
	%		{{\small RGW - ShapeDTW}}    
	%		\label{fig:MDS_14}
	%	\end{subfigure}
	\begin{subfigure}{0.3\textwidth}   
		\centering 
		\includegraphics[width=1.2\textwidth]{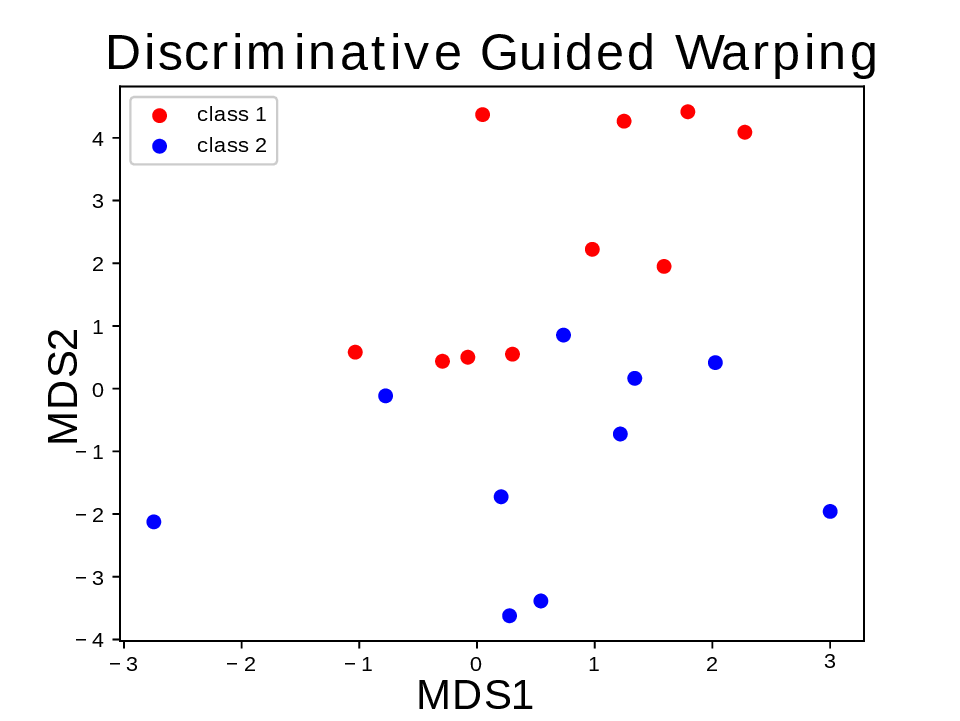}
		\caption[]%
		{{\small DGW}}    
		\label{fig:MDS_15}
	\end{subfigure}
	\\ 
	
	%	\begin{subfigure}{0.2\textwidth}   
	%		\centering 
	%		\includegraphics[width=1.2\textwidth]{discriminative_guided_warp_shape.eps}
	%		\caption[]%
	%		{{\small DGW - ShapeDTW}}    
	%		\label{fig:MDS_16}
	%	\end{subfigure}
	
	~~ 
	\begin{subfigure}{0.3\textwidth}   
		\centering 
		\includegraphics[width=1.2\textwidth]{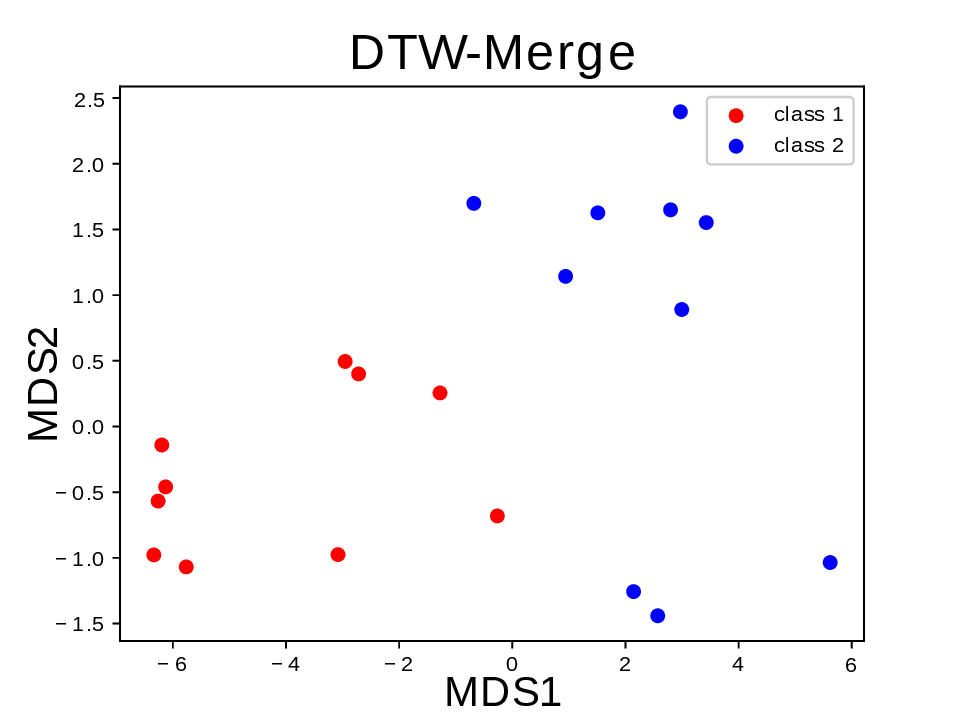}
		\caption[]%
		{{\small DTW-Merge}}    
		\label{fig:MDS_17}
	\end{subfigure}
	
	\caption[]
	{\small The MDS visualization of GAP layer for different data augmentation methods.} 
	\label{fig:MDS}
	
\end{figure*}

We also evaluated our technique using 1NN-DTW. Comparing the 1NN-DTW results with and without data augmentation shows a slight decrease of about $0.2$ percent, which means our method has no tangible effect on the DTW space of the time series. In Table \ref{table:types}, we categorized the datasets in the UCR archive according to their types. The 'Win' column shows the percentage of the datasets in which the ResNet with data augmentation performs better. As we can see, our method has the most positive effect on 'Hemodynamics,' 'Sensor,' 'Device,' 'Spectrum,' and 'Motion.'

\subsection{Grad-CAM}

In this section, we elaborate the utilization of grad-CAM for the ResNet trained by GunPoint and Strawberry datasets and depict that DTW-Merge helps the network to extract more discriminative and concentrated features.

The GunPoint dataset contains two classes: pointing with the index finger (class-1) and drawing a gun (class-2). These actions were performed by two actors, one male, and one female. The X-axis of the centroid of the right hand is recorded as a sequence and is used for the classification. The Strawberry dataset samples were obtained by Fourier-transform infrared spectroscopy (FTIR) and classified into two different classes, strawberry (class-1) and non-strawberry (class-2).   

After training the ResNet with and without data augmentation on the datasets mentioned above, we employed the grad-CAM technique to highlight the regions that play a significant part in classifying samples in a certain class for each dataset. The discriminative region between two classes of GunPoint and Strawberry datasets are shown in Fig~\ref{fig:grad_cam} and Fig~\ref{fig:grad_cam_2}, respectively. As we can see, when we use the data augmentation method, the network classifies samples by focusing on the discriminative region. Making decisions by emphasizing smaller parts of the time series could be more reliable since other unimportant parts might fool the classifier and cause misclassification.

\subsection{Comparison to other methods}

In Table \ref{table:comp}, we presented the comparison between data augmentation methods. other methods' results provided by \cite{Iwana_2020em}. The \textit{accuracy} column shows the average precision classification for all datasets, and the \textit{t} shows the t-test value between the corresponding method and when no augmentation was used. As we can see, DTW-Merge outperforms other methods.

In this part, we trained the ResNet with data generated by various data augmentation methods on 'BeetleFly' dataset. Then, we plotted the extracted features (GAP layer) for the test data using Multidimensional Scaling (MDS) to observe the performance of the trained network on extracting discriminative features. Metric MDS aims to preserve the actual Euclidean distance between the vectors by minimizing the \textit{Stress} loss function shown in Eq. \ref{eq:mds}.

\begin{table}[]
	\caption{The comparison of DTW-merge to other data augmentation methods. A negative t-value indicates the method deteriorated the neural network's performance compared to when no data augmentation is used.}

	\centering
	\resizebox{.8\textwidth}{!}{%
		\begin{tabular}{ccc}
			\multicolumn{1}{c|}{Data Augmentation Method}      & \multicolumn{1}{c|}{Accuracy (\%)} & t                    \\ \hline
			\multicolumn{1}{c|}{Jittering}                     & \multicolumn{1}{c|}{80.87±18.08}   & -0.87                \\
			\multicolumn{1}{c|}{Rotation}                      & \multicolumn{1}{c|}{78.01±19.93}   & -3.31                \\
			\multicolumn{1}{c|}{Scaling}                       & \multicolumn{1}{c|}{81.92±16.81}   & 1.05                 \\
			\multicolumn{1}{c|}{Magnitude Warping}             & \multicolumn{1}{c|}{81.33±17.15}   & -0.08                \\
			\multicolumn{1}{c|}{Permutation}                   & \multicolumn{1}{c|}{80.67±18.31}   & -1.16                \\
			\multicolumn{1}{c|}{Slicing}                       & \multicolumn{1}{c|}{81.51±17.64}   & 0.24                 \\
			\multicolumn{1}{c|}{Time Warping}                  & \multicolumn{1}{c|}{79.62±19.20}   & -2.25                \\
			\multicolumn{1}{c|}{Window Warping}                & \multicolumn{1}{c|}{82.32±16.93}   & 1.99                 \\
			\multicolumn{1}{c|}{SPAWNER}                       & \multicolumn{1}{c|}{80.47±17.02}   & -1.25                \\
			\multicolumn{1}{c|}{WDBA}                          & \multicolumn{1}{c|}{81.46±18.59}   & 0.12                 \\
			\multicolumn{1}{c|}{Random Guided Warping (RGW)}         & \multicolumn{1}{c|}{81.25±18.10}   & -0.22                \\
			\multicolumn{1}{c|}{Discriminative Guided Warping (DGW)} & \multicolumn{1}{c|}{81.99±17.38}   & 1.29                 \\
			\multicolumn{1}{c|}{\textbf{DTW-Merge}}                      & \multicolumn{1}{c|}{\textbf{83.07±16.25}}   & \textbf{ 2.1 }                 \\ \hline
			\multicolumn{1}{l}{}                               & \multicolumn{1}{l}{}               & \multicolumn{1}{l}{}
		\end{tabular}%
	}
	\label{table:comp}
\end{table}

\begin{equation}
Stress(x_{1}, x_{2}, ..., x_{N}) = (\sum_{i\ne j=1,...,N}{(d_{ij} - \lVert x_{i} - x_{j} \lVert})^2)^{1/2}
\label{eq:mds}
\end{equation}

In this equation, $d_{ij}$ is the Euclidean distance between the $i^{th}$ and $j^{th}$ feature vector and $x_{i}$ and $x_{j}$ are the points in MDS visualization (which in our case are 2D points).

In Fig. \ref{fig:MDS}, we depicted the MDS visualization of the GAP layer while feeding the 'BeetleFly' test set to ResNet trained by augmented data. As we can see, when we use the DTW-Merge, ResNet extracts more discriminative features, and the test data are linearly separable; however, it didn't happen when we used other methods.

\section{Conclusion}
\label{sec:conclusion}	

In this paper, a novel DTW-based data augmentation technique for time series is proposed named DTW-Merge. This method is based on DTW, which is a renowned tool for aligning time series. DTW-Merge uses this concept that the aligned parts of time series have the same temporal characteristics. This method enhances the performance of ResNet, a well-known neural network architecture for time series classification and, outperforms other data augmentation methods.

It is worth noting that we utilized ShapeDTW \cite{ZHAO_2018} to implement the our method, but we didn't see any improvement in results. Sometimes the samples of the same class have dissimilarities that cause an improper temporal alignment between sequences. For addressing this issue,  we can cluster the samples of each class and then select the samples from the same cluster for data augmentation. This solution will help the algorithm generate more realistic time series and be considered for future work.

\bibliographystyle{IEEEbib}
\bibliography{references.bib}

\end{document}